\documentclass[10pt,journal,compsoc]{IEEEtran}
\ifCLASSOPTIONcompsoc
  \usepackage[nocompress]{cite}
\else
  \usepackage{cite}
\fi
\ifCLASSINFOpdf
\usepackage[pdftex]{graphicx}
\else
\fi
\usepackage{booktabs}
\usepackage{subfigure}
\usepackage[ruled]{algorithm2e}
\usepackage{algpseudocode}
\usepackage{multirow}

\begin{document}
\title{Contrastive Clustering}

\author{Yunfan Li,
    Peng Hu,
    Zitao Liu,
    Dezhong Peng,
    Joey Tianyi Zhou,
    Xi Peng
% Michael~Shell,~\IEEEmembership{Member,~IEEE,}% <-this % stops a space
\IEEEcompsocitemizethanks{
\IEEEcompsocthanksitem Y. Li, P. Hu, D. Peng, and X. Peng are with the College of Computer Science, Sichuan University, 610065, Chengdu, China.
\IEEEcompsocthanksitem Z. Li is with AI Lab, TAL Education Group, 100080, Beijing, China.
\IEEEcompsocthanksitem J. Zhou is with IHPC, Institute for Infocomm Research, A*STAR, 138632, Singapore. 
\IEEEcompsocthanksitem Corresponding author: X. Peng.}
}
\IEEEtitleabstractindextext{%
\begin{abstract}
In this paper, we propose a one-stage online clustering method called Contrastive Clustering (CC) which explicitly performs the instance- and cluster-level contrastive learning. To be specific, for a given dataset, the positive and negative instance pairs are constructed through data augmentations and then projected into a feature space. Therein, the instance- and cluster-level contrastive learning are respectively conducted in the row and column space by maximizing the similarities of positive pairs while minimizing those of negative ones. Our key observation is that the rows of the feature matrix could be regarded as soft labels of instances, and accordingly the columns could be further regarded as cluster representations. By simultaneously optimizing the instance- and cluster-level contrastive loss, the model jointly learns representations and cluster assignments in an end-to-end manner. Extensive experimental results show that CC remarkably outperforms 17 competitive clustering methods on six challenging image benchmarks. In particular, CC achieves an NMI of 0.705 (0.431) on the CIFAR-10 (CIFAR-100) dataset, which is an up to 19\% (39\%) performance improvement compared with the best baseline.
\end{abstract}

}

\maketitle

\IEEEdisplaynontitleabstractindextext

% \fi
\IEEEpeerreviewmaketitle

\IEEEraisesectionheading{\section{Introduction}\label{sec:introduction}}

\IEEEPARstart{A}{s} one of the most fundamental tools in unsupervised learning, clustering could group data into different clusters without any label. Although some promising results have been achieved recently~\cite{liu2016multiple, liu2017sparse}, most of the algorithms would produce inferior results on complex datasets due to insufficient representability. To solve the problem, deep clustering~\cite{guo2017deep, ghasedi2017deep} utilizes neural networks to extract representative information from images for facilitating the downstream clustering tasks. In very recent, the focus of the community has shifted to how to learn representation and perform clustering in an end-to-end fashion. For example, JULE~\cite{JULE} progressively merges data points and takes the clustering results as supervisory signals to learn a more discriminative representation by a neural network. DeepClustering~\cite{DeepClustering} iteratively groups the features with k-means and uses the subsequent assignments to update the deep network. This kind of alternation-learning method would suffer from the error accumulated during the alternation between the stages of representation learning and clustering, which results in suboptimal clustering performance. Moreover, the aforementioned methods can only deal with offline tasks, \textit{i.e.}, the clustering is based on the whole dataset, which limits their application on large-scale online learning scenarios.

\begin{figure}[t]\centering
    \includegraphics[scale=0.27]{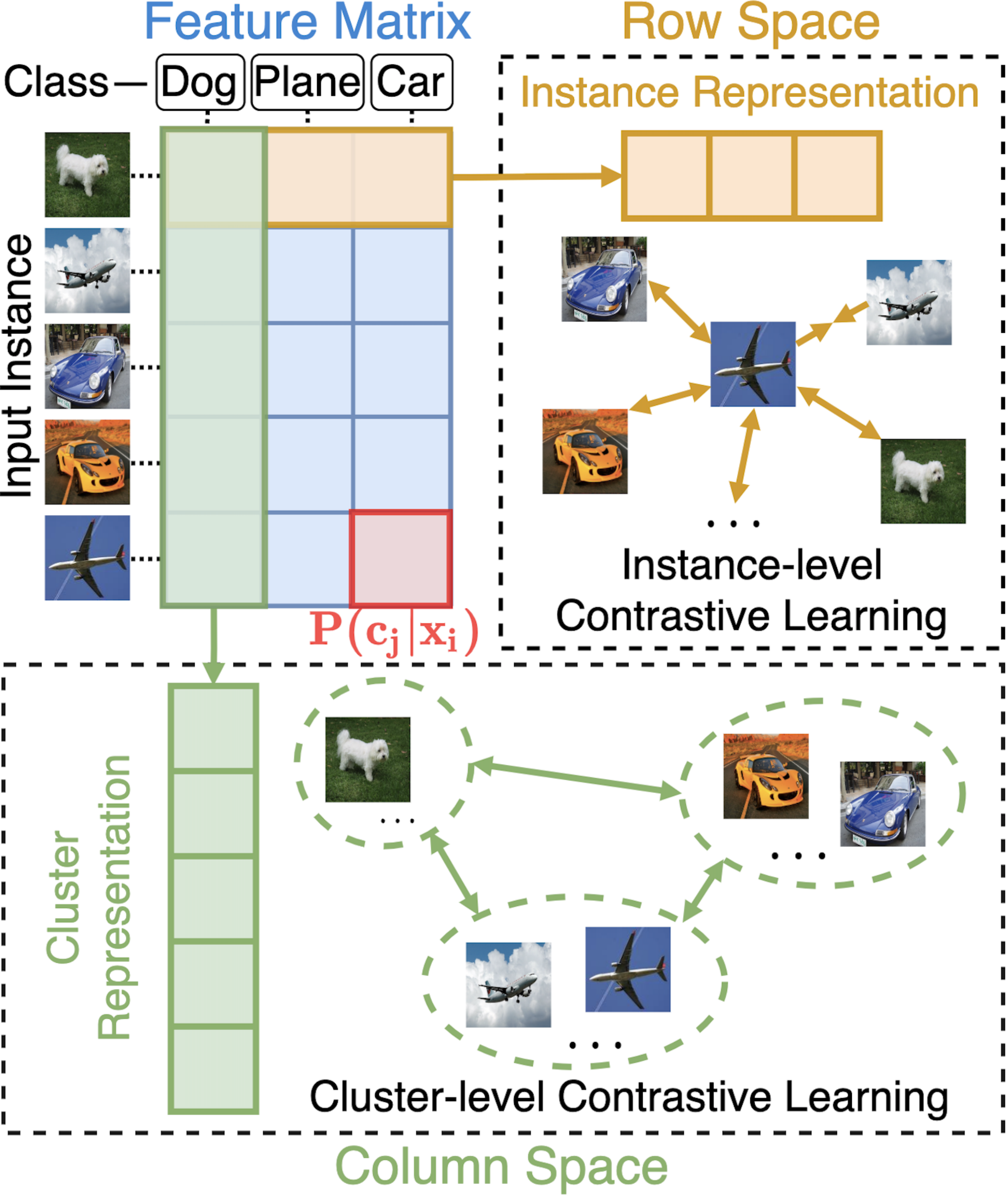}
    \caption{The key observation. By regarding the rows of the feature matrix as the soft labels of instances (\textit{i.e.}, $P(c_j|x_i)$ denotes the probability of sample $i$ belonging to cluster $j$), the columns could accordingly be interpreted as cluster representations distributed over the dataset. As a result, the instance- and cluster-level contrastive learning could be conducted in the row and column space of the feature matrix, respectively. }
    \label{fig:overview}
\end{figure}

To conquer the aforementioned offline limitation, this paper proposes a one-stage online deep clustering method called Contrastive Clustering (CC). Our idea comes from the observations shown in Fig.~\ref{fig:overview}. For a given dataset, we use a deep network to learn the feature matrix whose rows and columns correspond to the instance and cluster representations, respectively. In other words, we treat the label as a special representation by projecting input instances into a subspace with a dimensionality of the cluster number. In this sense, the rows of the feature matrix could be interpreted as the cluster assignment probabilities (\textit{i.e.}, instance soft labels), and the columns could then be regarded as the cluster distributions over instances (\textit{i.e.}, cluster representations). Owing to the observation of ``label as representation'', it is feasible to perform online clustering since the clustering prediction is now recast as a special representation learning task that is ``independent'' of other instances.

With the above observations, we propose a novel dual contrastive learning framework to learn instance and cluster representations. Specifically, CC first learns the feature matrix of data pairs constructed through a variety of data augmentations such as random crop and blurring. After that, the instance- and cluster-level contrastive learning are conducted in the row and column space of the feature matrix by gathering the positive pairs and scattering the negatives. By considering the instance- and cluster-level similarity under our dual contrastive learning framework, CC is able to simultaneously learn discriminative features and perform online clustering in a one-stage and end-to-end manner. To summarize, the major contributions of our work are as follows:
\begin{itemize}
	\item We provide a novel insight to the community, \textit{i.e.}, the instance representation and clustering prediction correspond to the row and column of a learnable feature matrix, respectively. Hence, deep clustering could be elegantly unified into the framework of representation learning; 
	\item To the best of our knowledge, this could be the first work of clustering-specified contrastive learning. Different from existing studies in contrastive learning, the proposed method conducts contrastive learning at not only the instance-level but also the cluster-level. Such a dual contrastive learning framework could produce clustering-favorite representations as proved in our experiments;
	\item The proposed model works in a one-stage and end-to-end fashion, which only needs batch-wise optimization and thus can be applied to large-scale online scenarios. 
\end{itemize}

The proposed method shows superior performance on six challenging image datasets, including CIFAR-10/100, STL-10, ImageNet-10/Dogs, and Tiny-ImageNet. It significantly outperforms state-of-the-art methods on all six datasets. In particular, it achieves an up to 39\% performance improvement in terms of NMI on the CIFAR-100 dataset compared with the most competitive baseline.

\section{Related Work}
\label{sec:2}
In this section, we briefly introduce some recent developments in two related topics, namely, contrastive learning and deep clustering.

\subsection{Contrastive Learning}
\label{sec:2.1}
As a promising paradigm of unsupervised learning, contrastive learning has lately achieved state-of-the-art performance in representation learning~\cite{BYOL}. The basic idea of contrastive learning is to map the original data to a feature space wherein the similarities of positive pairs are maximized while those of negative pairs are minimized~\cite{DimensionalityReductionbyLearninganInvariantMapping}. In early works, the positive and negative pairs are known as prior. Recently, various works have shown that large quantities of data pairs are crucial to the performance of contrastive models~\cite{MOCO} and they could be constructed using the following two strategies under the unsupervised setting. One is to use clustering results as pseudo labels to guide the pair construction~\cite{ClusteringBasedContrastiveLearning}. The other, which is more direct and commonly used, is to treat each instance as a class represented by a feature vector and data pairs are constructed through data augmentations~\cite{dosovitskiy2014discriminative}. To be specific, the positive pair composes of two augmented views of the same instance, and the other pairs are defined to be negative. Given the data pairs, several loss functions have been proposed for contrastive learning. For example, triplet loss~\cite{FaceNet} minimizes the distance between an anchor and a positive, while maximizing the distance between the anchor and a negative, NCE~\cite{NCE} performs nonlinear logistic regression to discriminate between the observed data and some artificially generated noise, and SimCLR~\cite{SimCLR} adopts the normalized temperature-scaled cross-entropy loss (NT-Xent) to identify positive pairs across the dataset.

The differences between our method and existing contrastive learning methods are addressed below. On the one hand, the existing works only perform contrastive learning at the instance level, whereas our method simultaneously conducts contrastive learning at both the instance- and cluster-level following the observation of ``label as representation''.  On the other hand, the existing works aim to learn a general representation, which is off-the-shelf for the downstream tasks. On the contrary, our method is specifically designed for clustering, which could be the first successful attempt of task-specified contrastive learning.
\begin{figure*}[t]\centering
    \includegraphics[scale=0.35]{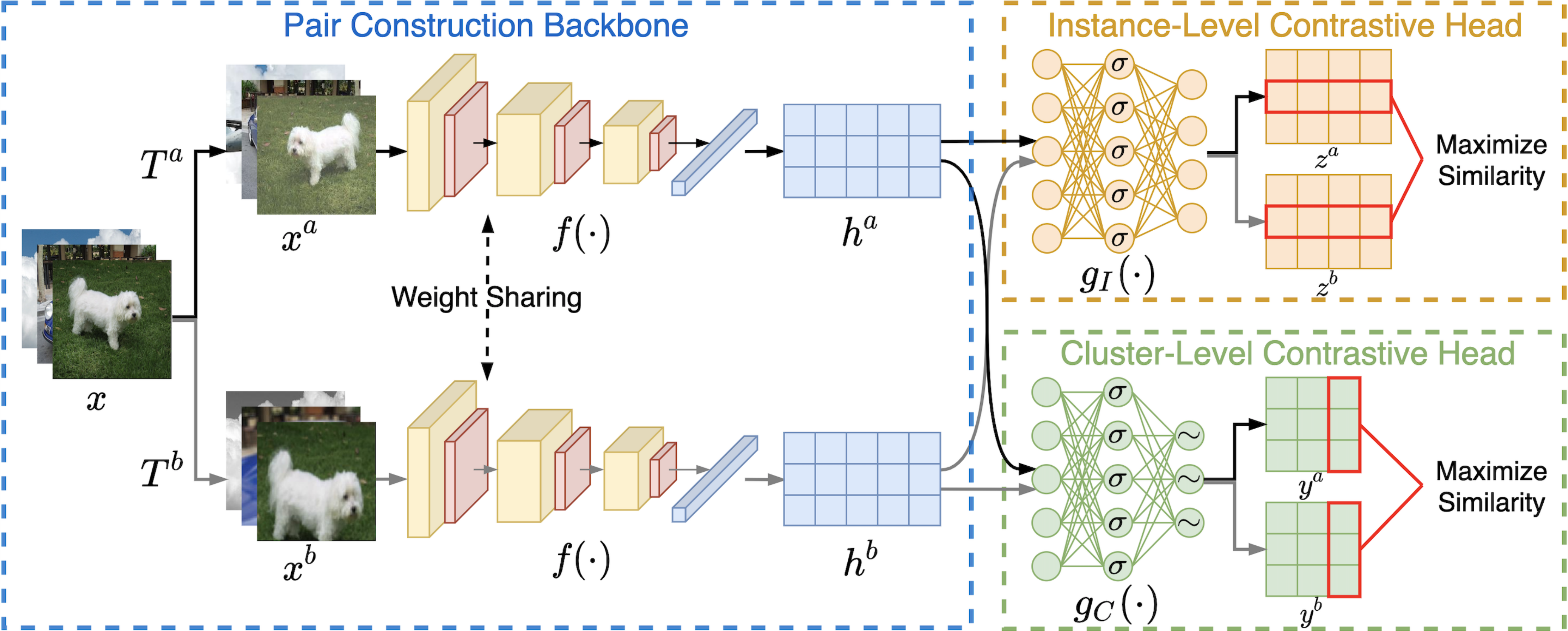}
    \caption{The framework of Contrastive Clustering. We construct data pairs using two data augmentations. Given data pairs, one shared deep neural network is used to extract features from different augmentations. Two separate MLPs ($\sigma$ denotes the \textit{ReLU} activation and $\sim$ denotes the \textit{Softmax} operation to produce soft labels) are used to project the features into the row and column space wherein the instance- and cluster-level contrastive learning are conducted respectively.}\label{fig:framework}
\end{figure*}
\subsection{Deep Clustering}
\label{sec:2.2}
Although promising results have been achieved, traditional clustering algorithms give discouraging results on large-scale complex datasets due to the inferior capability of representation learning. Benefit from the powerful representative ability of deep neural networks, deep clustering~\cite{DEC, deepclustering_wangqi} has shown promising performance on complex datasets. For example, JULE~\cite{JULE} performs agglomerative clustering by iteratively learning the data representations and cluster assignments. Analogously, DeepClustering~\cite{DeepClustering} groups the features using k-means and updates the deep network according to the cluster assignments in turn. Though this kind of two-stage methods could jointly learn representations and perform clustering, their performance might be hurt by the errors accumulated during the alternation. Besides, the entire dataset is needed to perform clustering, which limits their application in large-scale and online scenarios. Recently, some online clustering methods have been proposed~\cite{Peng2017:Cascade_full, IIC, PICA}. For example, IIC~\cite{IIC} discovers clusters by maximizing mutual information between the cluster assignments of data pairs. PICA~\cite{PICA} learns the most semantically plausible data separation by maximizing the partition confidence of the clustering solution. Though grounded in theory, these works rely heavily on the auxiliary over-clustering trick which is hard to explain.

Different from the above deep clustering methods, we treat the label as a special representation so that the instance- and cluster-level representation learning could be conducted in the row and column space, respectively. Besides, former works mainly utilize the representative capability of deep neural networks for clustering, whereas our method dually utilizes contrastive samples to facilitate clustering under a unified framework. Such a clustering-oriented contrastive learning paradigm helps the model to minimize the inter-cluster similarities to separate different clusters. To the best of our knowledge, this could be one of the first successful attempts to promote clustering through contrastive learning.

\section{Method}
\label{sec:3}
As illustrated in Fig.~\ref{fig:framework}, our method consists of three jointly learned components, namely, a pair construction backbone  (PCB), an instance-level contrastive head (ICH), and a cluster-level contrastive head (CCH). In brief, PCB constructs data pairs through data augmentations and extracts features from augmented samples, after that ICH and CCH respectively apply contrastive learning in the row and column space of the feature matrix. After training, the cluster assignments can be easily obtained through the soft labels predicted by CCH. Notably, although our basic idea indicates that the dual contrastive learning could be directly conducted on the feature matrix, we experimentally find that the clustering performance could be improved by decoupling the instance- and cluster-level contrastive learning into two independent subspaces  (see more details in the supplementary materials). The possible reason is that such a decoupling strategy could improve the representability of ICH and CCH. In the following, we will elaborate on the three components in turn and introduce the proposed objective function at the end.

\subsection{Pair Construction Backbone}
Inspired by the recent progress in contrastive learning~\cite{SimCLR}, CC uses data augmentations to construct data pairs. Specifically, given a data instance $x_i$, two stochastic data transformations $T^a, T^b$ sampled from the same family of augmentations $\mathcal{T}$ are applied to it, resulting in two correlated samples denoted as $x^a_i=T^a(x_i)$ and $x^b_i=T^b(x_i)$. The previous works have suggested that a proper choice of augmentation strategy is essential to achieve a good performance in downstream tasks. In this work, five types of data augmentation methods are used, including ResizedCrop, ColorJitter, Grayscale, HorizontalFlip, and GaussianBlur. For a given image, each augmentation is applied independently with a certain probability following the setting in SimCLR~\cite{SimCLR}. Specifically, ResizedCrop crops an image to a random size and resize the crop to the original size; ColorJitter changes the brightness, contrast, and saturation of an image; Grayscale converts an image to grayscale; HorizontalFlip horizontally flip an image and GaussianBlur blurs an image by a Gaussian function. 

One shared deep neural network $f(\cdot)$ is used to extracted features from the augmented samples via $h^a_i=f(x^a_i)$ and $h^b_i=f(x^b_i)$. As for the architecture of the network, theoretically, our method does not depend on a specific network. Here, we simply adopt ResNet34~\cite{ResNet} as the backbone for fair comparison.

\subsection{Instance-level Contrastive Head}

\begin{algorithm}[t]
\caption{Contrastive Clustering}
\label{algorithm}
\KwIn{dataset $\mathcal{X}$; training epochs $E$; batch size $N$; temperature parameter $\tau_I$ and $\tau_C$; cluster number $M$; structure of $\mathcal{T}$, $f$, $g_I$, and $g_C$.}

\KwOut{cluster assignments.}

\tcp{training}

\For{epoch = 1 to E}{
	sample a mini-batch $\{x_i\}_{i=1}^N$ from $\mathcal{X}$
		
	sample two augmentations $T^a, T^b \sim \mathcal{T}$
	
	compute instance and cluster representations by
	$h^a_i = f(T^a(x_i)), h^b_i = f(T^b(x_i))$
	$z^a_i = g_I(h^a_i), z^b_i = g_I(h^b_i)$
	$\tilde{y}^a_i=g_C(h^a_i), \tilde{y}^b_i=g_C(h^b_i)$
	
	compute instance-level contrastive loss $\mathcal{L}_{ins}$ through Eq.~\ref{eq:instance similarity}--\ref{eq:instance loss}
	
	compute cluster-level contrastive loss $\mathcal{L}_{clu}$ through Eq.~\ref{eq:cluster similarity}--\ref{eq:cluster loss}
	
	compute overall loss $\mathcal{L}$ by Eq.~\ref{eq:overall loss}
	
	update $f, g_I, g_C$ to minimize $\mathcal{L}$
}

\tcp{test}

\For{$x$ in $\mathcal{X}$}{
	extract features by $h = f(x)$
	
	compute cluster assignment by $c=\arg\max g_C(h)$
}
\end{algorithm}

Contrastive learning aims to maximize the similarities of positive pairs while minimizing those of negative ones. The characteristics of pairs can be defined by different criteria. For example, one can define pairs of within-class samples to be positive and leave the others negative. In this work, since no prior label is available on the clustering task, the positive and negative pairs are constructed at the instance-level according to pseudo-labels generated by data augmentations. More specifically, the positive pairs consist of samples augmented from the same instance, and the negative pairs otherwise. 

Formally, given a mini-batch of size $N$, CC performs two types of data augmentations on each instance $x_i$ and results in $2N$ data samples $\{x^a_1, \dots , x^a_N, x^b_1, \dots , x^b_N\}$. For a specific sample $x^a_i$, there are $2N-1$ pairs in total, among which we choose its corresponding augmented sample $x^b_i$ to form a positive pair $\{x^a_i, x^b_i\}$ and leave other $2N-2$ pairs to be negative.

To alleviate the information loss induced by contrastive loss, we do not directly conduct contrastive learning on the feature matrix. Instead, we stack a two-layer nonlinear MLP $g_I(\cdot)$ to map the feature matrix to a subspace via $z^a_i=g_I(h^a_i)$ where the instance-level contrastive loss is applied. The pair-wise similarity is measured by cosine distance, \textit{i.e.}, 
\begin{equation}
\label{eq:instance similarity}
  s(z^{k_1}_i, z^{k_2}_j) = \frac{(z^{k_1}_i)(z^{k_2}_j)^\top}{\|z^{k_1}_i\|\|z^{k_2}_j\|},
\end{equation}
where $k_1, k_2 \in \{a, b\}$ and $i, j \in [1, N]$. To optimize pair-wise similarities, without loss of generality, the loss for a given sample $x^a_i$ is in the form of
\begin{equation}
  \ell^a_i = -\log \frac{\exp(s(z^a_i, z^b_i)/\tau_I)}{\sum_{j=1}^N [\exp(s(z^a_i, z^a_j)/\tau_I) + \exp(s(z^a_i, z^b_j)/\tau_I)]},
\end{equation}
where $\tau_I$ is the instance-level temperature parameter. Since we hope to identify all positive pairs across the dataset, the instance-level contrastive loss is computed over every augmented samples, namely, 
\begin{equation}
\label{eq:instance loss}
  \mathcal{L}_{ins} = \frac{1}{2N} \sum_{i=1}^N (\ell^a_i + \ell^b_i).
\end{equation}

\subsection{Cluster-level Contrastive Head}

Following the idea of ``label as representation", when projecting a data sample into a space whose dimensionality equals to the number of clusters, the $i$-th element of its feature can be interpreted as its probability of belonging to the $i$-th cluster, and the feature vector denotes its soft label accordingly.

Formally, let $Y^a\in \mathcal{R}^{N\times M}$ be the output of CCH for a mini-batch under the first augmentation (and $Y^b$ for the second augmentation), and then $Y^a_{n,m}$ can be interpreted as the probability of sample $n$ being assigned to cluster $m$, where $N$ is the batch size and $M$ equals to the number of clusters. Since each sample belongs to only one cluster, ideally, the rows of $Y^a$ tends to be one-hot. In this sense, the $i$-th column of $Y^a$ can be seen as a representation of the $i$-th cluster and all columns should differ from each other. 

Similar to $g_I(\cdot)$ used in the instance-level contrastive head, we use another two-layer MLP $g_C(\cdot)$ to project the feature matrix into an $M$-dimensional space via $\tilde{y}^a_i=g_C(h^a_i)$, where $\tilde{y}^a_i$ denotes the soft label of sample $x^a_i$ (the $i$-th row of $Y^a$). For clarity, let $y^a_i$ be the $i$-the column of $Y^a$, namely, the representation of cluster $i$ under the first data augmentation, and we combine it with $y^b_i$ to form a positive cluster pair $\{y^a_i, y^b_i\}$, while leaving other $2M-2$ pairs to be negative, where $y^b_i$ denotes the second augmented representation of cluster $i$. Again, we use cosine distance to measure the similarity between cluster pairs, that is
\begin{equation}
\label{eq:cluster similarity}
  s(y^{k_1}_i, y^{k_2}_j) = \frac{(y^{k_1}_i)^\top(y^{k_2}_j)}{\|y^{k_1}_i\|\|y^{k_2}_j\|},
\end{equation}
where $k_1, k_2 \in \{a, b\}$ and $i, j \in [1, M]$. Without loss of generality, the following loss function is adopted to distinguish cluster $y^a_i$ from all other clusters except $y^b_i$, \textit{i.e.},
\begin{equation}
  {\hat{\ell}}^a_i = -\log \frac{\exp(s(y^a_i, y^b_i)/\tau_C)}{\sum_{j=1}^M [\exp(s(y^a_i, y^a_j)/\tau_C) + \exp(s(y^a_i, y^b_j)/\tau_C)]},
\end{equation}
where $\tau_C$ is the cluster-level temperature parameter. By traversing all clusters, the cluster-level contrastive loss is finally computed by
\begin{equation}
\label{eq:cluster loss}
  \mathcal{L}_{clu} = \frac{1}{2M} \sum_{i=1}^M (\hat{\ell}^a_i + \hat{\ell}^b_i) - H(Y),
\end{equation}
where $H(Y)=\sum_{i=1}^M [P(y^a_i)\log P(y^a_i)+P(y^b_i)\log P(y^b_i)]$ is the entropy of cluster assignment probabilities $P(y^k_i) = \sum_{j=1}^N Y^k_{ji}\ /\ \|Y\|_{1}, k\in \{a, b\}$ within a mini-batch under each data augmentation. This term helps to avoid the trivial solution that most instances are assigned to the same cluster.

\subsection{Objective Function}

The optimization of ICH and CCH is a one-stage and end-to-end process. Two heads are simultaneously optimized and the overall objective function consists of the instance-level and cluster-level contrastive loss, \textit{i.e.},
\begin{equation}
\label{eq:overall loss}
  \mathcal{L} = \mathcal{L}_{ins} + \mathcal{L}_{clu}.
\end{equation}

Generally, a dynamic weight parameter could be applied to balance the two losses across the training process~\cite{BYOL}, but in practice, we find a simple addition of the two losses already works well. The full training and test process of the model is summarized in Algorithm~\ref{algorithm}.

\section{Experiments}
In this section, we conduct experiments to verify the effectiveness of the proposed CC. 

\subsection{Experimental Configurations}
We first introduce the used datasets, implementations, and the used performance metrics. 
\subsubsection{Datasets}

We evaluate the proposed method on six challenging image datasets. A brief description of these datasets is summarized in Table~\ref{tab:datasets}. Both the training and test set are used for CIFAR-10, CIFAR-100~\cite{CIFAR}, and STL-10~\cite{STL}, while only the training set is used for ImageNet-10, ImageNet-Dogs~\cite{ImageNet10/dogs}, and Tiny-ImageNet~\cite{TinyImageNet}. For CIFAR-100, its 20 super-classes rather than 100 classes are taken as the ground-truth. For STL-10, its 100,000 unlabeled samples are additionally used to train the instance-level contrastive head. 

\begin{table}[h]
\centering
\caption{A summary of datasets used for evaluations.}
\begin{tabular}{@{}cccc@{}}
\toprule
Dataset       & Split      & Samples & Classes \\ \midrule
CIFAR-10      & Train+Test & 60,000   & 10      \\
CIFAR-100     & Train+Test & 60,000   & 20      \\
STL-10        & Train+Test & 13,000   & 10      \\
ImageNet-10   & Train      & 13,000   & 10      \\
ImageNet-Dogs & Train      & 19,500   & 15      \\
Tiny-ImageNet & Train      & 100,000  & 200     \\ \bottomrule
\end{tabular}

\label{tab:datasets}
\end{table}

\subsubsection{Implementation Details}

For a fair comparison with previous works~\cite{IIC, PICA}, we adopt ResNet34 as the backbone network. As ResNet is designed for images of size $224 \times 224$, some previous works modified the standard ResNet and used some tricks (\textit{e.g.}, the Sobel layer used in PICA) to help the network to handle small-sized inputs (\textit{e.g.}, CIFAR-10). However, these specialized modifications and tricks should vary with images of different sizes, which brings difficulty in model selection. In this work, we simply resize all input images to the size of $224 \times 224$, and no modification is applied to the standard ResNet. Notably, as up-scaling already leads to blurred images, we leave the GaussianBlur augmentation out for the small image collections including CIFAR-10, CIFAR-100, STL-10, and Tiny-ImageNet.

For the instance-level contrastive head, the dimensionality of the row space is set to 128 to keep more information of images, and the instance-level temperature parameter $\tau_I$ is fixed to $0.5$ in all experiments. For the choice of the dimensionality of the row space, we conduct additional analysis in the supplementary material. As for the cluster-level contrastive head, the dimensionality of the column space is naturally set to the number of clusters, and the cluster-level temperature parameter $\tau_C = 1.0$ is used for all datasets.

The Adam optimizer with an initial learning rate of 0.0003 is adopted to simultaneously optimize the two contrastive heads and the backbone network. No weight decay or scheduler is used. The batch size is set to 256 due to the memory limitation, and we train the model from scratch for 1,000 epochs to compensate for the performance loss caused by small batch size as suggested in SimCLR~\cite{SimCLR}. The experiments are carried out on Nvidia TITAN RTX 24G and it takes about 70 gpu-hours to train the model on CIFAR-10, 90 gpu-hours for CIFAR-100, 160 gpu-hours on STL-10, 20 gpu-hours on ImageNet-10, 30 gpu-hours on ImageNet-dogs, and 130 gpu-hours on Tiny-ImageNet.

\subsubsection{Evaluation Metrics}

Three widely-used clustering metrics including Normalized Mutual Information (NMI), Accuracy (ACC), and Adjusted Rand Index (ARI) are utilized to evaluate our method. Higher values of these metrics indicate better clustering performance. 

\subsection{Comparisons with State of the Arts}
We evaluate the proposed CC on six challenging image benchmarks and compare it with 17 representative state-of-the-art clustering approaches, including k-means~\cite{Kmeans}, SC~\cite{SC}, AC~\cite{AC}, NMF~\cite{NMF}, AE~\cite{AE}, DAE~\cite{DAE}, DCGAN~\cite{DCGAN}, DeCNN~\cite{DeCNN}, VAE~\cite{VAE}, JULE~\cite{JULE}, DEC~\cite{DEC}, DAC~\cite{DAC}, ADC~\cite{ADC}, DDC~\cite{DDC}, DCCM~\cite{DCCM}, IIC~\cite{IIC} and PICA~\cite{PICA}. For SC, NMF, AE, DAE, DCGAN, DeCNN, and VAE, clustering results are obtained via k-means on the features extracted from images. 

\begin{table*}[t]
\centering
\caption{The clustering performance on six challenging object image benchmarks. The best results are shown in boldface.}
\resizebox{\textwidth}{!}{
\begin{tabular}{@{}lcccccccccccccccccc@{}}
\toprule
Dataset &
  \multicolumn{3}{c}{CIFAR-10} &
  \multicolumn{3}{c}{CIFAR-100} &
  \multicolumn{3}{c}{STL-10} &
  \multicolumn{3}{c}{ImageNet-10} &
  \multicolumn{3}{c}{ImageNet-Dogs} &
  \multicolumn{3}{c}{Tiny-ImageNet} \\ \midrule
Metrics &
  \multicolumn{1}{l}{NMI} &
  \multicolumn{1}{l}{ACC} &
  \multicolumn{1}{l}{ARI} &
  \multicolumn{1}{l}{NMI} &
  \multicolumn{1}{l}{ACC} &
  \multicolumn{1}{l}{ARI} &
  \multicolumn{1}{l}{NMI} &
  \multicolumn{1}{l}{ACC} &
  \multicolumn{1}{l}{ARI} &
  \multicolumn{1}{l}{NMI} &
  \multicolumn{1}{l}{ACC} &
  \multicolumn{1}{l}{ARI} &
  \multicolumn{1}{l}{NMI} &
  \multicolumn{1}{l}{ACC} &
  \multicolumn{1}{l}{ARI} &
  \multicolumn{1}{l}{NMI} &
  \multicolumn{1}{l}{ACC} &
  \multicolumn{1}{l}{ARI} \\ \midrule
K-means    & 0.087 & 0.229 & 0.049 & 0.084 & 0.130 & 0.028 & 0.125 & 0.192 & 0.061 & 0.119 & 0.241 & 0.057 & 0.055 & 0.105 & 0.020 & 0.065 & 0.025 & 0.005 \\
SC         & 0.103 & 0.247 & 0.085 & 0.090 & 0.136 & 0.022 & 0.098 & 0.159 & 0.048 & 0.151 & 0.274 & 0.076 & 0.038 & 0.111 & 0.013 & 0.063 & 0.022 & 0.004 \\
AC         & 0.105 & 0.228 & 0.065 & 0.098 & 0.138 & 0.034 & 0.239 & 0.332 & 0.140 & 0.138 & 0.242 & 0.067 & 0.037 & 0.139 & 0.021 & 0.069 & 0.027 & 0.005 \\
NMF        & 0.081 & 0.190 & 0.034 & 0.079 & 0.118 & 0.026 & 0.096 & 0.180 & 0.046 & 0.132 & 0.230 & 0.065 & 0.044 & 0.118 & 0.016 & 0.072 & 0.029 & 0.005 \\
AE         & 0.239 & 0.314 & 0.169 & 0.100 & 0.165 & 0.048 & 0.250 & 0.303 & 0.161 & 0.210 & 0.317 & 0.152 & 0.104 & 0.185 & 0.073 & 0.131 & 0.041 & 0.007 \\
DAE        & 0.251 & 0.297 & 0.163 & 0.111 & 0.151 & 0.046 & 0.224 & 0.302 & 0.152 & 0.206 & 0.304 & 0.138 & 0.104 & 0.190 & 0.078 & 0.127 & 0.039 & 0.007 \\
DCGAN      & 0.265 & 0.315 & 0.176 & 0.120 & 0.151 & 0.045 & 0.210 & 0.298 & 0.139 & 0.225 & 0.346 & 0.157 & 0.121 & 0.174 & 0.078 & 0.135 & 0.041 & 0.007 \\
DeCNN      & 0.240 & 0.282 & 0.174 & 0.092 & 0.133 & 0.038 & 0.227 & 0.299 & 0.162 & 0.186 & 0.313 & 0.142 & 0.098 & 0.175 & 0.073 & 0.111 & 0.035 & 0.006 \\
VAE        & 0.245 & 0.291 & 0.167 & 0.108 & 0.152 & 0.040 & 0.200 & 0.282 & 0.146 & 0.193 & 0.334 & 0.168 & 0.107 & 0.179 & 0.079 & 0.113 & 0.036 & 0.006 \\
JULE       & 0.192 & 0.272 & 0.138 & 0.103 & 0.137 & 0.033 & 0.182 & 0.277 & 0.164 & 0.175 & 0.300 & 0.138 & 0.054 & 0.138 & 0.028 & 0.102 & 0.033 & 0.006 \\
DEC        & 0.257 & 0.301 & 0.161 & 0.136 & 0.185 & 0.050 & 0.276 & 0.359 & 0.186 & 0.282 & 0.381 & 0.203 & 0.122 & 0.195 & 0.079 & 0.115 & 0.037 & 0.007 \\
DAC        & 0.396 & 0.522 & 0.306 & 0.185 & 0.238 & 0.088 & 0.366 & 0.470 & 0.257 & 0.394 & 0.527 & 0.302 & 0.219 & 0.275 & 0.111 & 0.190 & 0.066 & 0.017 \\
ADC        & -     & 0.325 & -     & -     & 0.160 & -     & -     & 0.530 & -     & -     & -     & -     & -     & -     & -     & -     & -     & -     \\
DDC        & 0.424 & 0.524 & 0.329 & -     & -     & -     & 0.371 & 0.489 & 0.267 & 0.433 & 0.577 & 0.345 & -     & -     & -     & -     & -     & -     \\
DCCM       & 0.496 & 0.623 & 0.408 & 0.285 & 0.327 & 0.173 & 0.376 & 0.482 & 0.262 & 0.608 & 0.710 & 0.555 & 0.321 & 0.383 & 0.182 & 0.224 & 0.108 & 0.038 \\
IIC        & -     & 0.617 & -     & -     & 0.257 & -     & -     & 0.610 & -     & -     & -     & -     & -     & -     & -     & -     & -     & -     \\
PICA & 0.591 & 0.696 & 0.512 & 0.310 & 0.337 & 0.171 & 0.611 & 0.713 & 0.531 & 0.802 & 0.870 & 0.761 & 0.352 & 0.352 & 0.201 & 0.277 & 0.098 & 0.040 \\
\textbf{CC(Ours)} & \textbf{0.705} & \textbf{0.790} & \textbf{0.637} & \textbf{0.431} & \textbf{0.429} & \textbf{0.266} & \textbf{0.764} & \textbf{0.850} & \textbf{0.726} & \textbf{0.859} & \textbf{0.893} & \textbf{0.822} & \textbf{0.445} & \textbf{0.429} & \textbf{0.274} & \textbf{0.340} & \textbf{0.140} & \textbf{0.071} \\ \bottomrule
\end{tabular}
}
\label{tab:result}
\end{table*}

According to the results shown in Table~\ref{tab:result}, CC significantly outperforms these state-of-the-art baselines by a large margin on all six datasets. In particular, CC surpasses the closest competitor PICA by 0.114 on CIFAR-10, 0.121 on CIFAR-100, and 0.153 on STL-10 in terms of NMI. Moreover, CC achieves more than 50\% performance improvements on the best baseline on CIFAR-100 and Tiny-ImageNet in terms of ARI. The remarkable results demonstrate the powerful clustering ability of CC, which benefits from the incorporation of the instance- and cluster-level contrastive learning.

\begin{figure}[t]\centering
    \includegraphics[scale=0.22]{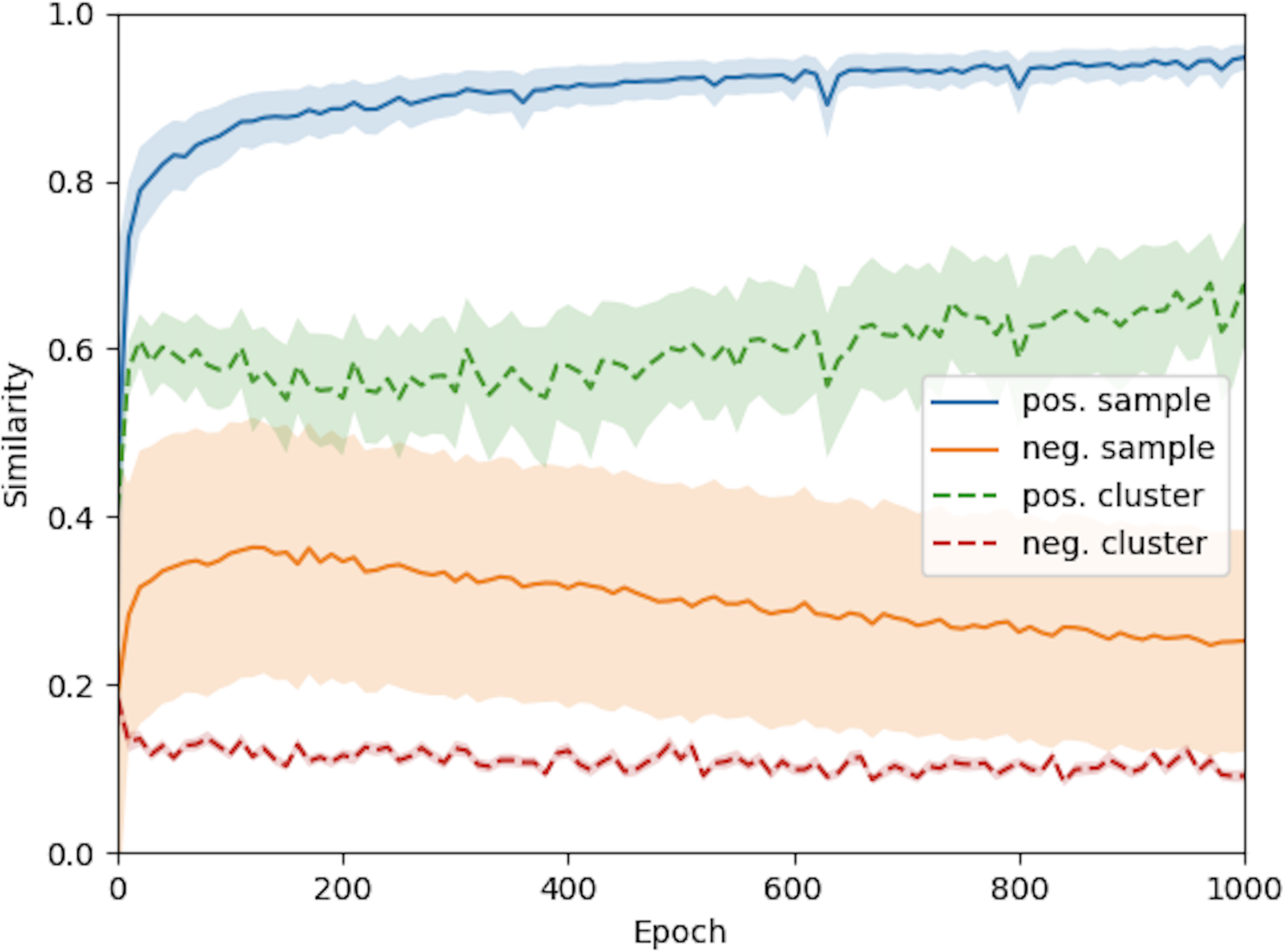}
    \caption{Instance-level and cluster-level pair-wise similarities across the training process on ImageNet-10. The colored areas denote the variances.}\label{fig:similarity}
\end{figure}

\begin{figure*}[h]\centering
  \subfigure[0 epoch (NMI = 0.183)]{
    \includegraphics[scale=0.312]{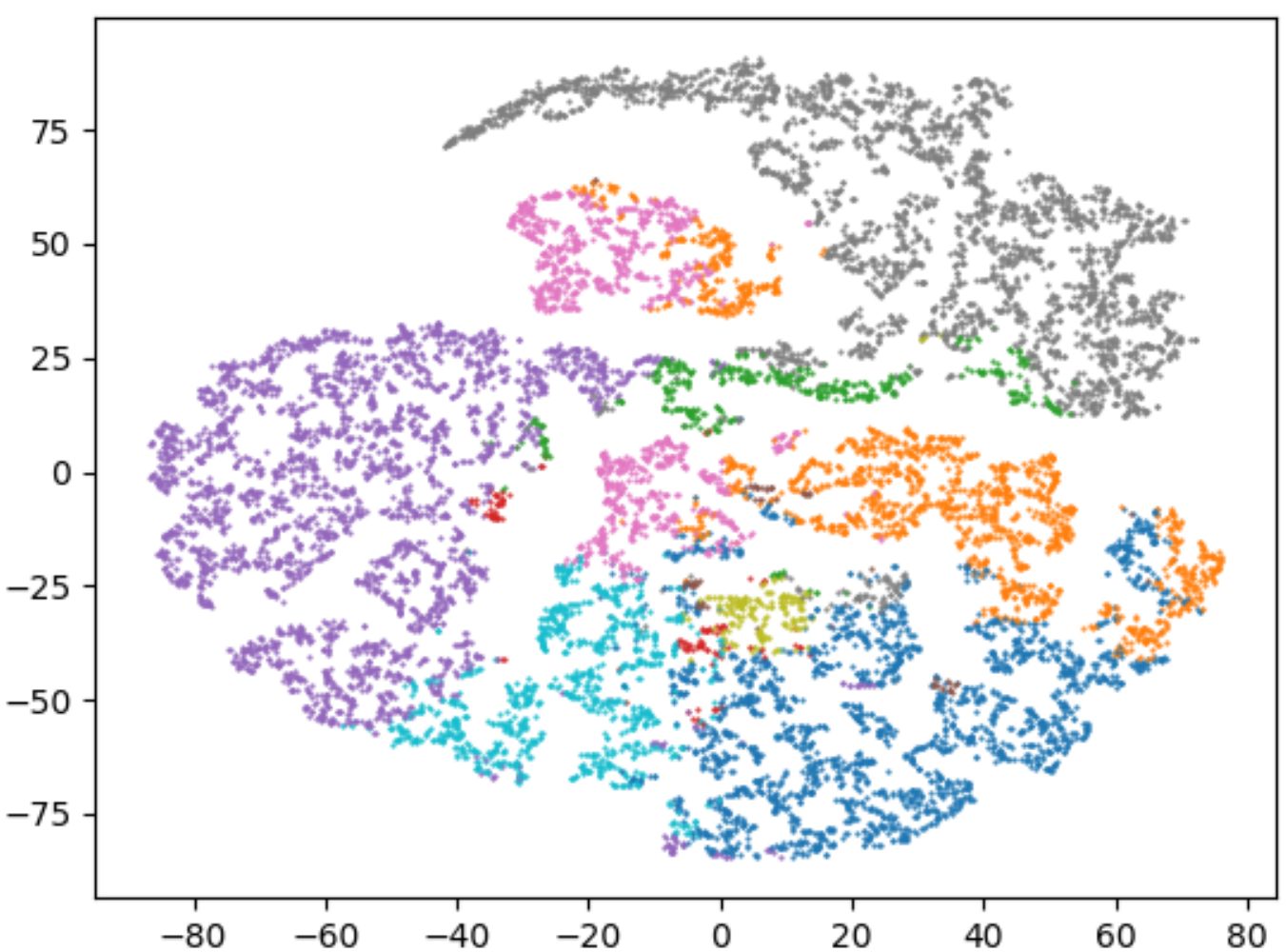}}
  \subfigure[20 epoch (NMI = 0.472)]{
    \includegraphics[scale=0.312]{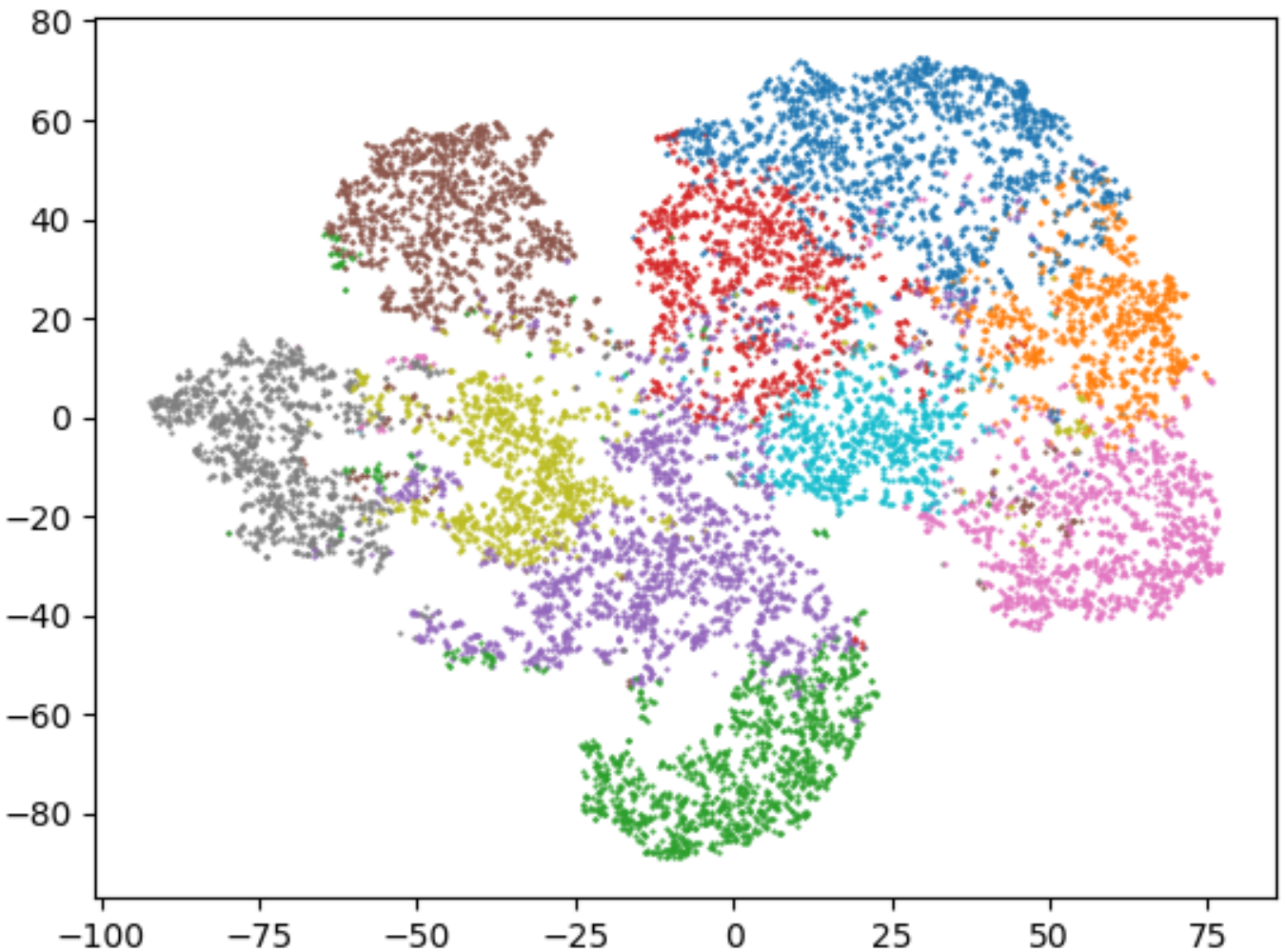}}
  \subfigure[50 epoch (NMI = 0.628)]{
    \includegraphics[scale=0.312]{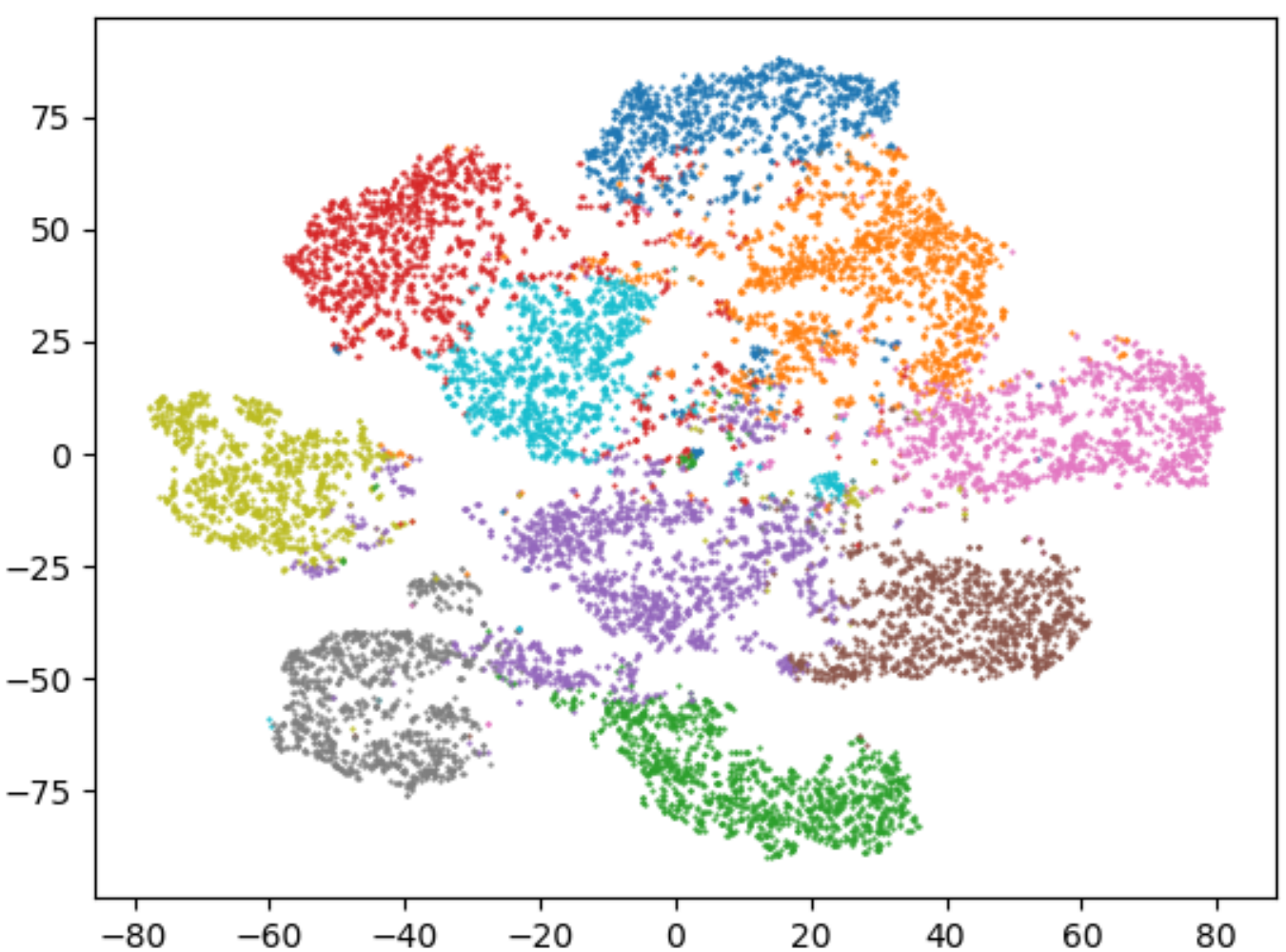}}
  \subfigure[100 epoch (NMI = 0.737)]{
    \includegraphics[scale=0.312]{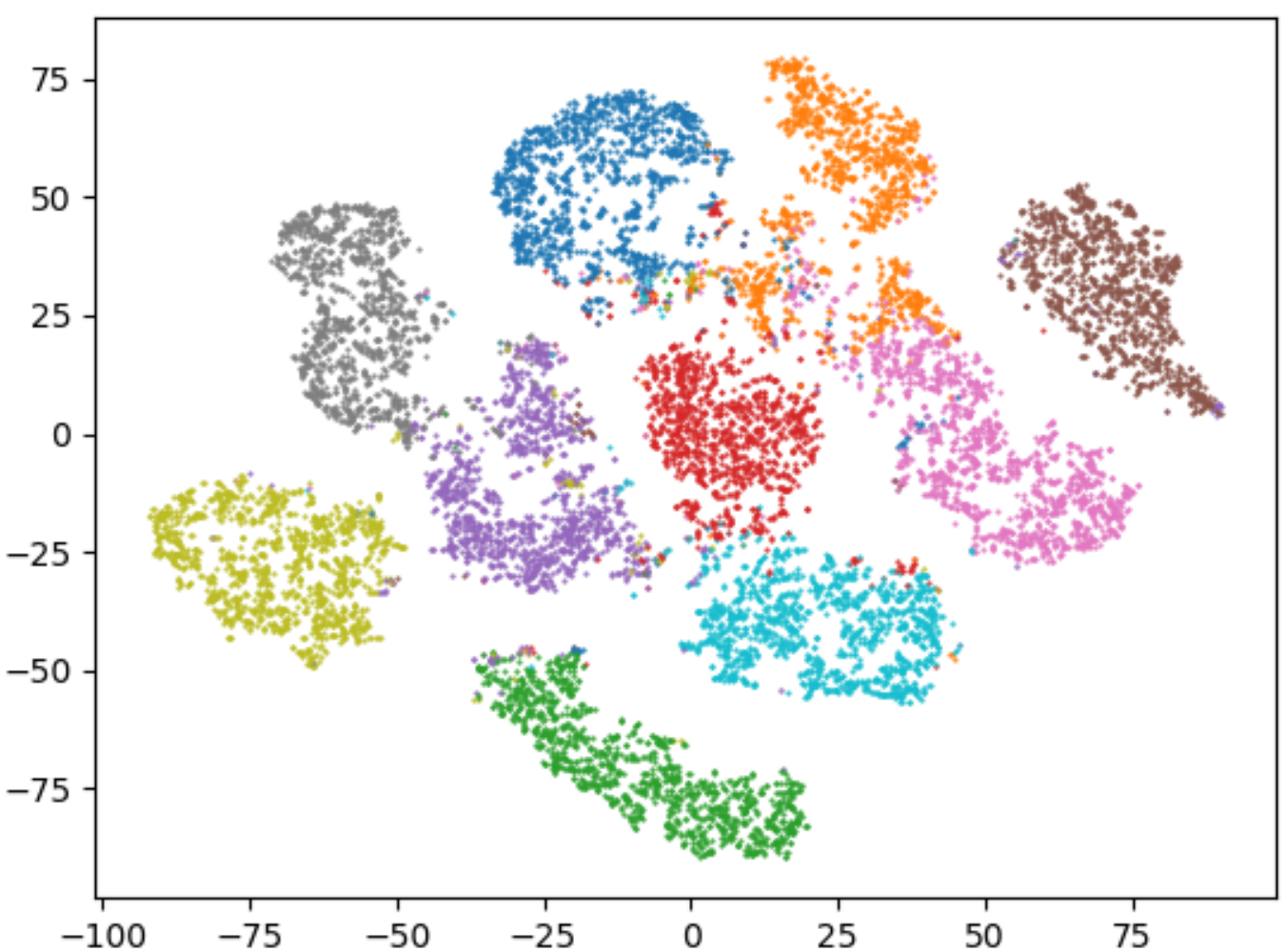}}
  \caption{The evolution of instance features and cluster assignments across the training process on ImageNet-10. The colors indicate the cluster assignment obtained from CCH and the features for t-SNE are computed from ICH.}\label{fig:tsne}
\end{figure*}

\subsection{Qualitative Study}

We carry out two experiments to analyze the pair-wise similarity across the training process and the evolution of the learned instance representation and cluster assignments on ImageNet-10.

\subsubsection{Analysis on Pair-wise Similarity}

To provide an intuitive understanding of how contrastive clustering works, we visualize the changes of both the instance- and cluster-level pair-wise similarities w.r.t. the training epoch. As shown in Fig.~\ref{fig:similarity}, the similarities of positive instance/cluster pairs grows as the training process goes while the similarity of negative instance/cluster pairs stay at a low level. In addition, the similarity interval between the positive and negative is comparatively large at both the instance- and cluster-level, which explains the success of our model. Note that the variances of positive instance and negative cluster pairs are much lower than those of negative instance pairs and positive cluster pairs due to the following two reasons. On the one hand, the large variance of negative instance pairs could be attributed to the fact that some pairs consist of samples of different instances but the same class, which should be treated as positive theoretically. On the other hand, the variance of positive cluster pairs comes from the inconsistent cluster assignments of samples under different augmentations.

\subsubsection{Evolution of Instance Feature and Cluster Assignments}

By simultaneously optimizing the instance- and cluster-level contrastive head, the model ought to learn discriminative representations and desirable cluster assignments at the same time. To see how our model converges to the goal, we perform t-SNE in the row space at four different timestamps throughout the training process. The results are shown in Fig.~\ref{fig:tsne}, where different colors indicated different labels predicted by the cluster-level contrastive head. The result shows that, at the beginning, features are all mixed and most instances are assigned to a few clusters. As the training process goes, cluster assignments become more reasonable, and features scatter and gather more distinctly.

\subsection{Ablation Study}

Three ablation studies are carried out to further understand the importance of data augmentation, the effect of two contrastive heads, and the reliance on the backbone network.

\subsubsection{Importance of Data Augmentation}
Some existing works have shown that the performance of contrastive learning heavily relies on the proper strategy of data augmentation~\cite{SimCLR}. To verify the significance of data augmentation, we test our model on CIFAR-10 and ImageNet-10 by removing one and both of the two augmentations. When data augmentations are removed, the raw image is directly used as the input. Table~\ref{tab:augmentation} shows that data augmentations could enhance the performance of CC, especially on more complicated datasets, \textit{i.e.}, CIFAR-10. When no data augmentation is applied,  every positive pair consists of two same samples/clusters and thus only negative pairs take part in model optimization, which leads to pretty poor results.

\begin{table}[h]
\centering
\caption{Importance of data augmentation.}
\begin{tabular}{@{}ccccc@{}}
\toprule
Dataset                   & Augmentation & NMI   & ACC   & ARI   \\ \midrule
\multirow{3}{*}{CIFAR-10} & $T^a(x) + T^b(x)$  & \textbf{0.705} & \textbf{0.790} & \textbf{0.637} \\
                          & $T^a(x) + x~~~~~~~~$     & 0.630 & 0.690 & 0.533 \\
                          & $x + x$          & 0.045 & 0.169 & 0.022 \\ \midrule
\multirow{3}{*}{ImageNet-10} & $T^a(x) + T^b(x)$ & \textbf{0.859} & \textbf{0.893} & \textbf{0.822} \\
                          & $T^a(x) + x~~~~~~~~$     & 0.852 & 0.892 & 0.817 \\
                          & $x + x$          & 0.063 & 0.177 & 0.030 \\ \bottomrule
\end{tabular}

\label{tab:augmentation}
\end{table}

\subsubsection{Effect of Contrastive Head}
To prove the effectiveness of the instance- and cluster-level contrastive head, we conduct ablation studies on CIFAR-10 and ImageNet-10 by removing one of the two heads. Since the cluster assignments can no longer be directly obtained when the cluster-level contrastive head is removed, we perform k-means in the instance space instead. The results are shown in Table~\ref{tab:head}. Interestingly, ICH shows comparable performance on CIFAR-10 while CCH performs better on ImageNet-10, which suggests the joint effects of the two heads to some extent.

\begin{table}[h]
\centering
\caption{Effect of two contrastive heads.}
\begin{tabular}{@{}ccccc@{}}
\toprule
Dataset                      & Contrastive Head & NMI            & ACC            & ARI            \\ \midrule
\multirow{3}{*}{CIFAR-10}    & ICH + CCH     & \textbf{0.705}          & \textbf{0.790}          & \textbf{0.637}          \\
                             & ICH Only      & 0.699 & 0.782 & 0.616 \\
                             & CCH Only      & 0.592 & 0.657 & 0.499 \\ \midrule
\multirow{3}{*}{ImageNet-10} & ICH + CCH     & \textbf{0.859} & \textbf{0.893} & \textbf{0.822} \\
                             & ICH Only      & 0.838          & 0.888          & 0.780          \\
                             & CCH Only      & 0.850          & 0.892          & 0.816          \\ \bottomrule
\end{tabular}

\label{tab:head}
\end{table}

\subsubsection{Reliance on Backbone Network}
In our framework, any feature extractor could be adopted as the backbone network. To examine how much our model relies on the structure of the backbone network, we test three ResNets of different depths and report the results in Table~\ref{tab:backbone}. The results suggest that the representability of the backbone network contributes to the clustering performance. On relatively simple datasets like ImageNet-10, ResNet18 is sufficiently powerful to extract discriminative features. In addition, the performance of ResNet50 is worse than ResNet34 on CIFAR-10, which suggests a deeper network does not promise better performance.

\begin{table}[h]
\centering
\caption{Reliance on backbone network.}
\begin{tabular}{@{}ccccc@{}}
\toprule
Dataset                      & Backbone & NMI & ACC & ARI \\ \midrule
\multirow{3}{*}{CIFAR-10}    & ResNet18 & 0.650 & 0.736 &  0.569 \\
                             & ResNet34 & \textbf{0.705} & \textbf{0.790} & \textbf{0.637} \\
                             & ResNet50 & 0.663 & 0.747 & 0.585 \\ \midrule
\multirow{3}{*}{ImageNet-10} & ResNet18 & 0.851 & 0.889 & 0.816 \\
                             & ResNet34 & 0.859 & 0.893 & 0.822 \\
                             & ResNet50 & \textbf{0.859} & \textbf{0.895} & \textbf{0.823} \\\bottomrule
\end{tabular}

\label{tab:backbone}
\end{table}

%\section{Discussion}
%
%\subsubsection{Rethinking IIC from Contrastive Learning Perspective}
%In this section, we provide an interpretation of the important over-clustering trick used in IIC from the contrastive learning perspective. 
%\begin{equation}
%  derivation\ from\ IIC\ loss\ to\ contrastive\ loss
%\end{equation}
%
%[Optimizing mutual information leads to the NCE contrastive loss function]
%
%[Using more negative pairs increases the lower bound of mutual information]
%
%[The over-clustering trick used by IIC can be seen as performing contrastive learning in a higher dimensional space, where more feature information is kept]

\section{Conclusion}
Based on the observation that the rows and columns of the feature matrix could be respectively realized as the representation of instances and clusters, we proposed the Contrastive Clustering (CC) method which dually conducts contrastive learning at the instance- and cluster-level under a unified framework. The proposed CC shows its promising performance in clustering. In the future, we plan to extend it to other tasks and applications such as semi-supervised learning and transfer learning.

\ifCLASSOPTIONcaptionsoff
  \newpage
\fi



\bibliographystyle{IEEEtran}



%\end{IEEEbiography}

%\newpage



%\vfill


\end{document}